\documentclass{article}
\usepackage{iclr2026_conference,times}

\usepackage{amsmath,amsfonts,bm}

\def\eqref#1{equation~\ref{#1}}

\def\1{\bm{1}}

\DeclareMathAlphabet{\mathsfit}{\encodingdefault}{\sfdefault}{m}{sl}
\SetMathAlphabet{\mathsfit}{bold}{\encodingdefault}{\sfdefault}{bx}{n}

\usepackage{hyperref}
\usepackage{url}
\usepackage{booktabs}
\usepackage{amsfonts}
\usepackage{amsmath}
\usepackage{nicefrac}
\usepackage{microtype}
\usepackage{graphicx}
\usepackage{xcolor}
\usepackage{subcaption}
\usepackage{multirow}

\title{What Do World Models Learn in RL? \\ Probing Latent Representations in Learned Environment Simulators}

\iclrfinalcopy

\author{
Xinyu Zhang \\
Anyscale \\
\texttt{xinyu@gmail.com} \\
}

\newcommand{\Rsq}{R^2}

\begin{document}

\maketitle

\begin{abstract}
World models learn to simulate environment dynamics from experience, enabling sample-efficient reinforcement learning.
But what do these models actually represent internally?
We apply interpretability techniques---including linear and nonlinear probing, causal interventions, and attention analysis---to two architecturally distinct world models: IRIS (discrete token transformer) and DIAMOND (continuous diffusion UNet), trained on Atari Breakout and Pong.
Using linear probes, we find that both models develop \emph{linearly decodable} representations of game state variables (object positions, scores), with MLP probes yielding only marginally higher $\Rsq$, confirming that these representations are approximately linear.
Causal interventions---shifting hidden states along probe-derived directions---produce correlated changes in model predictions, providing evidence that representations are functionally used rather than merely correlated.
Analysis of IRIS attention heads reveals spatial specialization: specific heads attend preferentially to tokens overlapping with game objects.
Multi-baseline token ablation experiments consistently identify object-containing tokens as disproportionately important.
Our findings provide interpretability evidence that learned world models develop structured, approximately linear internal representations of environment state across two games and two architectures.
\end{abstract}

\section{Introduction}

World models---learned simulators of environment dynamics---have become a cornerstone of sample-efficient reinforcement learning \citep{ha2018world, hafner2023mastering}.
Recent advances such as IRIS \citep{micheli2023transformers} and DIAMOND \citep{alonso2024diffusion} achieve strong performance on the Atari 100k benchmark by learning to predict future observations entirely from experience.
Yet despite their empirical success, a fundamental question remains: \emph{what do world models represent internally?}

This question connects to the broader ``linear representation hypothesis'' emerging in mechanistic interpretability research \citep{park2023linear, elhage2022toy}.
\citet{li2023emergent} showed that a transformer trained to predict Othello moves develops an emergent linear representation of the board state, despite never being trained on it directly.
\citet{nanda2023emergent} extended this to chess.
If sequence models trained on game transcripts develop world representations, do world models---which are \emph{explicitly} trained to predict future observations---develop even richer ones?

We apply probing classifiers \citep{alain2017understanding, belinkov2022probing} and causal interventions to two architecturally distinct world models---IRIS \citep{micheli2023transformers} (VQ-VAE + transformer) and DIAMOND \citep{alonso2024diffusion} (UNet diffusion)---on Breakout and Pong.
We contribute: (1) linear and MLP probing showing approximately linear game state representations ($\Delta \leq 0.06$);
(2) causal interventions confirming representations are functionally used ($r > 0.95$);
(3) per-head spatial attention specialization; and
(4) multi-baseline token ablation with consistent importance rankings ($\rho > 0.9$).

\section{Method}

\subsection{Models and Ground Truth}

\textbf{IRIS} tokenizes $64 \times 64$ observations into 16 discrete tokens (VQ-VAE, codebook 512, $4 \times 4$ grid) then predicts sequences with a GPT-2 transformer (10 layers, 4 heads, dim 256).
\textbf{DIAMOND} uses a UNet denoiser (4 stages, 64 channels) with EDM preconditioning.
We probe on \textbf{Breakout} (\texttt{ball\_x}, \texttt{ball\_y}, \texttt{player\_x}, score) and \textbf{Pong} (\texttt{ball\_x}, \texttt{ball\_y}, \texttt{player\_y}, \texttt{enemy\_y}), with ground truth from Atari RAM \citep{anand2019unsupervised}.

\subsection{Probing Protocol}

We extract frozen representations from all layers ($N{=}10{,}000$ frames per game): IRIS VQ-VAE encoder/embedding + 10 transformer layers; DIAMOND conv input, 4 encoder/decoder stages, bottleneck, norm output.
For each (layer, property) pair, we train Ridge regression ($\alpha{=}1.0$) and 2-layer MLP probes (256$\to$128$\to$1, ReLU, Adam), both with 5-fold CV $\Rsq$.
The \emph{selectivity gap} $\Delta = \Rsq_\text{MLP} - \Rsq_\text{linear}$ measures nonlinear structure.
Controls: raw pixels, random model, shuffled labels.

\subsection{Causal Intervention Protocol}

To move beyond correlation, we perform activation patching along probe directions \citep{geiger2021causal}: modify hidden states $\mathbf{h}' = \mathbf{h} + \alpha \cdot \hat{\mathbf{w}}$ (where $\hat{\mathbf{w}}$ is the normalized Ridge probe weight) and measure the resulting change in next-token logits (KL divergence, token change rate).
A positive correlation between $|\alpha|$ and prediction change indicates the probe direction is \emph{functionally used}.

\subsection{Attention Analysis and Token Ablation}

For IRIS's 40 attention heads, we compute attention entropy and per-head spatial selectivity (mean attention to each of 16 token positions, forming $4\times4$ maps).
For token ablation, we replace each spatial token under three baselines (zero, mean, random codebook entry) and measure prediction disruption (KL divergence).
Consistency across baselines (Spearman $\rho$) indicates robust importance rankings.

\section{Results}

\subsection{Linear Representations Across Games}

\begin{figure}[t]
    \centering
    \includegraphics[width=\linewidth]{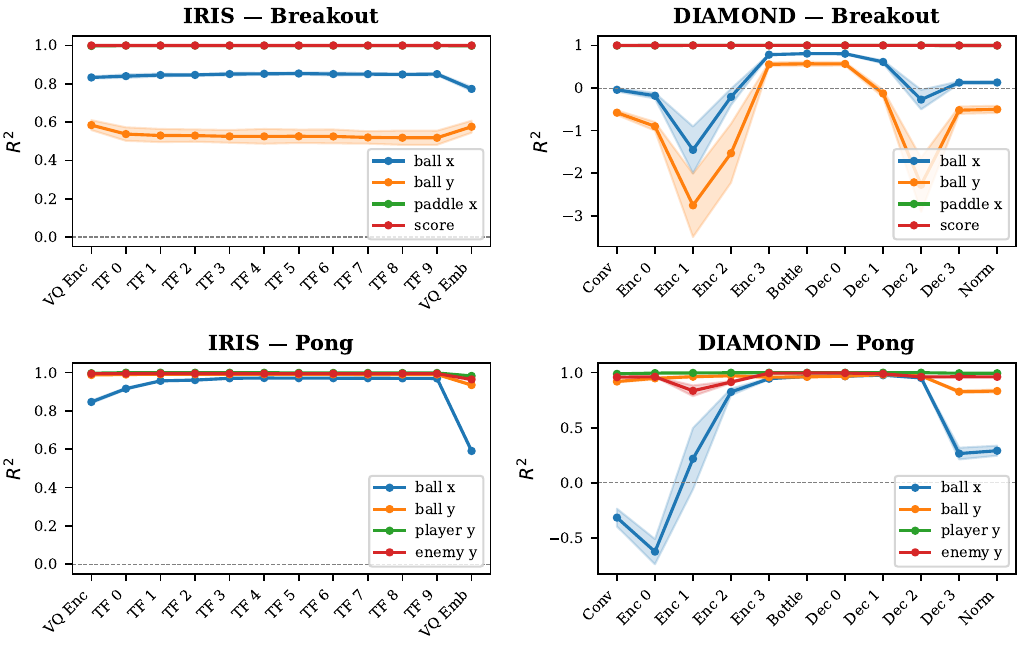}
    \caption{Probe $\Rsq$ across layers (in network data-flow order) for IRIS (left) and DIAMOND (right) on Breakout (top) and Pong (bottom). Each line tracks one game-state property; shaded bands show $\pm$1 std over 5-fold CV. IRIS representations are flat across transformer layers, while DIAMOND shows a peaked inverted-V centered on the UNet bottleneck. Note: $y$-axis includes negative $\Rsq$ values, revealing that DIAMOND's early encoder layers are worse than a constant predictor for ball position.}
    \label{fig:r2_lineplot}
\end{figure}

Figure~\ref{fig:r2_lineplot} shows probe $\Rsq$ across all layers for both games.

\paragraph{IRIS.}
Ball position encoding is stable across all transformer layers in both games: for Breakout, $\Rsq$ for \texttt{ball\_x} ranges from $0.84 \pm 0.01$ (layer~0) to $0.85 \pm 0.006$ (layer~5), a span of only $0.01$.
Paddle position and score are near-perfectly encoded ($\Rsq > 0.99$) at every layer.
The VQ-VAE encoder already achieves $\Rsq = 0.83$ for ball-x, and the transformer barely improves this ($+0.02$).

\paragraph{DIAMOND.}
Early encoder stages (0--2) show negative $\Rsq$ for ball position (as low as $-1.45$), while the deepest encoder stage begins recovering ($\Rsq = 0.78$); the bottleneck achieves peak linear $\Rsq = 0.81 \pm 0.01$ for \texttt{ball\_x}, and decoder stages decline---an inverted-V pattern suggesting the bottleneck compresses information maximally.
MLP probes recover substantially more from decoder layers ($\Rsq = 0.91$ at dec\_2 vs.\ $-0.27$ linear), indicating these layers encode ball position nonlinearly via skip connections.

\paragraph{Cross-game consistency.}
Pong shows the same architectural signatures: flat IRIS profiles and peaked DIAMOND bottleneck (Figure~\ref{fig:r2_lineplot}, bottom row), indicating these patterns are architecture-dependent rather than game-specific.
However, both models track ball position significantly better in Pong ($\Rsq \geq 0.95$) than Breakout ($\Rsq \leq 0.85$), likely because Pong's visual scene is simpler (no bricks, fewer objects).
Interestingly, DIAMOND's V-shape in ball tracking is much less pronounced in Pong (dec\_2 $\Rsq = 0.95$ vs.\ $-0.27$ in Breakout), suggesting this pattern is game-dependent.

\begin{table}[t]
\centering
\caption{Best-layer $\Rsq$ (mean $\pm$ std over 5-fold CV) for Breakout. Both linear and MLP probes are shown; the small selectivity gap ($\Delta$) confirms approximately linear representations.}
\label{tab:breakout}
\small
\begin{tabular}{lcccc}
\toprule
\textbf{Representation} & \texttt{ball\_x} & \texttt{ball\_y} & \texttt{player\_x} & \texttt{score} \\
\midrule
Random model & $-1.21$ & $-1.22$ & $-1.14$ & $-1.18$ \\
Shuffled labels & $-0.51$ & $-0.49$ & $-0.53$ & $-0.52$ \\
Raw pixels & $-1.31$ & $-0.48$ & $0.9989 \pm .0006$ & $0.9998 \pm .0001$ \\
\midrule
IRIS (Linear) & $\mathbf{0.85} \pm .006$ & $0.58 \pm .03$ & $\mathbf{0.9994} \pm .0001$ & $1.0000 \pm .0000$ \\
IRIS (MLP) & $0.91 \pm .005$ & $0.59 \pm .03$ & $0.9987 \pm .0002$ & $0.9999 \pm .0000$ \\
$\Delta_\text{IRIS}$ & $+0.06$ & $+0.01$ & $-0.0007$ & $-0.0001$ \\
\midrule
DIAMOND (Linear) & $0.81 \pm .01$ & $\mathbf{0.57} \pm .05$ & $1.0000 \pm .0000$ & $\mathbf{1.0000} \pm .0000$ \\
DIAMOND (MLP) & $0.91 \pm .005$ & $0.63 \pm .05$ & $0.9994 \pm .0002$ & $0.9998 \pm .0001$ \\
$\Delta_\text{DIAMOND}$ & $+0.10$ & $+0.06$ & $-0.0006$ & $-0.0002$ \\
\bottomrule
\end{tabular}
\vspace{-2mm}
\end{table}

Table~\ref{tab:breakout} shows that MLP probes yield only marginal improvements over linear probes for IRIS ($|\Delta| \leq 0.06$).
DIAMOND shows a larger gap for ball position ($\Delta = 0.10$ for \texttt{ball\_x}), driven by decoder layers where skip connections mix information nonlinearly; at the bottleneck itself, the gap is small ($\Delta = 0.04$).
Both models dramatically outperform baselines, with raw pixels failing on ball position ($\Rsq = -1.31$), showing that world model representations extract non-trivial structure.

For Pong, both models achieve even higher $\Rsq$: IRIS tracks ball position with $\Rsq = 0.97$ (linear) and $0.995$ (MLP); DIAMOND achieves $\Rsq = 0.98$ (linear) and $0.99$ (MLP).
Selectivity gaps are uniformly small ($\Delta \leq 0.03$), confirming approximately linear encoding across both games.

\subsection{Causal Interventions Confirm Functional Use}

\begin{figure}[t]
    \centering
    \includegraphics[width=0.75\linewidth]{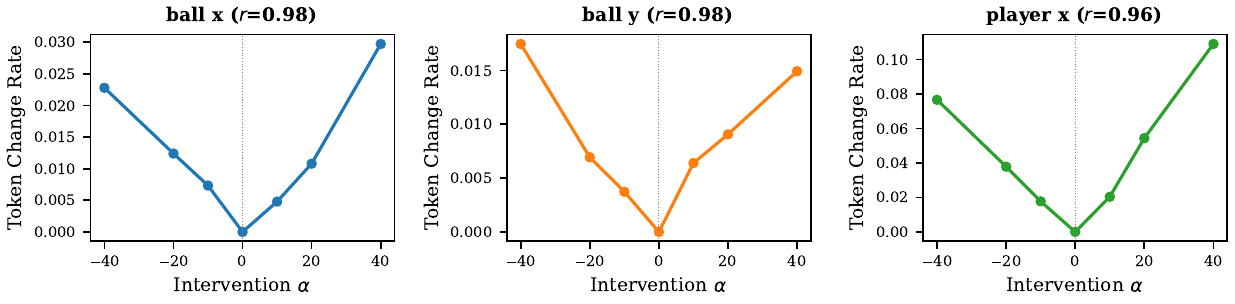}
    \vspace{-3mm}
    \caption{Causal intervention on Breakout: shifting IRIS layer-5 hidden states along probe directions produces correlated changes in predictions ($r \geq 0.96$ for all properties, measured via KL divergence).}
    \label{fig:causal}
\end{figure}

Figure~\ref{fig:causal} shows that shifting IRIS layer-5 hidden states along probe directions produces monotonically increasing prediction changes.
Correlation between $|\alpha|$ and KL divergence is strong: $r = 0.97$ (\texttt{ball\_x}), $r = 0.97$ (\texttt{ball\_y}), $r = 0.97$ (\texttt{player\_x}).
\texttt{player\_x} interventions produce ${\sim}16\times$ larger KL than ball interventions ($0.033$ vs.\ $0.002$ at $\alpha{=}40$), suggesting the model relies more on paddle position.
This confirms that linear representations are \emph{functionally used}, not mere artifacts.

\subsection{Attention and Token Ablation}

\begin{figure}[t]
    \centering
    \includegraphics[width=0.95\linewidth]{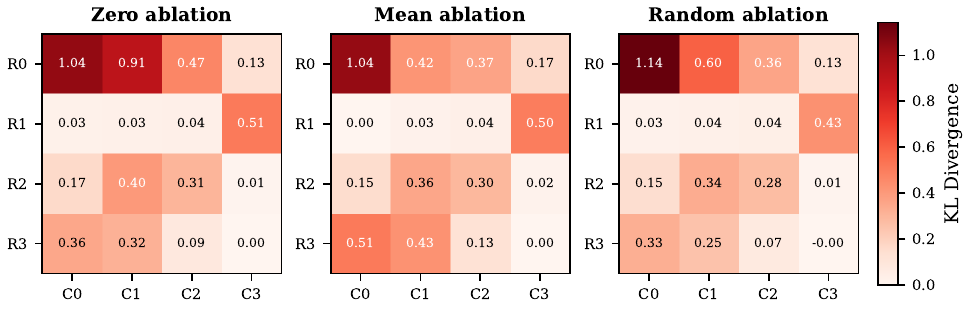}
    \vspace{-3mm}
    \caption{Three-way token ablation on Breakout ($4\times4$ grid). Zero, mean, and random replacement produce consistent importance rankings ($\rho > 0.92$), with token~0 (score/brick region) most critical.}
    \label{fig:ablation}
\end{figure}

Attention entropy ranges from $1.0$--$1.75$ nats across 40 heads (below $H_\text{max} = 2.83$), and individual heads show distinct spatial preferences: the four most selective heads---$(0,3)$, $(4,2)$, $(6,0)$, $(5,0)$---concentrate attention on different spatial regions, suggesting division of labor for tracking game elements.

Token ablation (Figure~\ref{fig:ablation}) consistently identifies token~0 (score/brick region) as most critical (KL $> 1.0$, ${\sim}50\%$ token change rate).
Rank correlation across methods is high ($\rho = 0.93$ zero/mean, $\rho > 0.99$ zero/random).
KL divergence correlates moderately with ball distance ($r \approx 0.56$ for Breakout), while Pong shows weaker spatial correlation ($r \approx 0.13$), suggesting information is distributed less spatially in simpler scenes.

\section{Discussion and Conclusion}

Our results demonstrate that learned world models develop structured, approximately linear internal representations of game state across two games and two architectures.
This parallels findings from the Othello-GPT line of work \citep{li2023emergent, nanda2023emergent}, extending them to pixel-based environment simulation.

\paragraph{Architectural comparison.}
IRIS's VQ-VAE tokenizer already produces strong linear representations ($\Rsq = 0.83$ for ball position), which the transformer preserves but barely improves ($\Rsq = 0.85$, a gain of only $+0.02$).
Rather than a limitation, this reveals a meaningful division of labor: the tokenizer handles spatial encoding while the transformer focuses on temporal dynamics and prediction---a factorization that single-frame probes cannot fully evaluate.
DIAMOND concentrates abstract state sharply at the UNet bottleneck; we hypothesize skip connections allow low-level information to bypass it, so only the bottleneck must encode abstract state.

\paragraph{Limitations.}
Both games are 2D Atari; generalization to 3D environments remains open.
Single-frame probes may miss temporal structure that the transformer encodes; sequence-conditioned probes could reveal richer dynamics.
Activation patching along a single direction is a coarse intervention; more targeted causal methods \citep{geiger2021causal} could strengthen these findings.

\bibliography{references}
\bibliographystyle{iclr2026_conference}

\end{document}